\documentclass[runningheads]{llncs}
\usepackage{amsmath}
\usepackage[T1]{fontenc}
\usepackage[misc]{ifsym}
\usepackage[export]{adjustbox}
\usepackage{amsmath, amssymb}
\usepackage{booktabs}
\usepackage{multirow}

\usepackage{enumitem}
\usepackage{url}
\usepackage{graphicx}
\usepackage{svg}
\usepackage{algorithm}
\usepackage{algorithmic}
\usepackage[most]{tcolorbox}   
\newcommand{\pmm}[1]{\,{\scriptsize$\pm$#1}}   

\begin{document}
\pagenumbering{gobble} 
%
%

\title{C-Unseen: Weak Signal Detection in Dynamic Temporal Knowledge Graphs via LLM Reasoning}

\titlerunning{Weak Signal Detection in DTKGs via LLM Reasoning.}

\author{Yassir LAIRGI\inst{1,2}\textsuperscript{\Letter}\and
Ludovic MONCLA\inst{1} \and
Khalid BENABDESLEM\inst{1}\and 
Rémy CAZABET\inst{1} \and
Pierre CLÉAU\inst{2}}
\authorrunning{Y. Lairgi et al.}
%
\institute{INSA Lyon, CNRS, UCBL, LIRIS, UMR5205, 69621 Villeurbanne, France \email{\{ludovic.moncla, khalid.benabdeslem, remy.cazabet\}@liris.cnrs.fr} \and
GAUC, Lyon, France \email{\{yassir.lairgi, pierre.cleau\}@auvalie.com}}

\maketitle              
\begin{abstract}

Weak signals are early, low-visibility indicators that precede significant changes before those changes become established. Existing detection methods, based on keyword frequency, topic modeling, or untyped graph topology, fail to capture the semantic and relational structure through which such signals manifest. In this paper, we propose \texttt{C-Unseen}\footnote{The code, prompts, and dataset are available at \url{https://github.com/AuvaLab/itext2kg}}, a self-interpretable framework for weak signal detection in Dynamic Temporal Knowledge Graphs (DTKGs). We define a weak signal as a rare, semantically coherent subgraph that proliferates across consecutive TKG snapshots. The framework operates through two modules: a Rare Subgraphs Extractor, in which an LLM identifies subgraphs whose content is in tension with the dominant snapshot narrative via chain-of-thought reasoning, and a Weak Signal Alerter, in which the persistence of these rare subgraphs is tracked across time steps to isolate true weak signals. Experimental results demonstrate that \texttt{C-Unseen} outperforms keyword-, topic-, and graph-based baselines.

\end{abstract}
\section{Introduction}
\label{sec:introduction}
In strategic domains, decisions must often be made before clear evidence emerges. The concept of weak signals addresses this challenge~\cite{ansoff1975managing}. A strong signal is unambiguous evidence that a change has already taken place: its effects are visible, and its consequences can be assessed with confidence. A weak signal is its precursor, appearing while the change is only beginning to take shape and too fragmentary on its own to prove that anything is happening at all. The example in Figure~\ref{fig:motivating-example} illustrates this distinction. Weak signals are characterized across three measurable dimensions: visibility, degree of emergence, and interpretability~\cite{hiltunen2008future}. Their significance, however, does not exist in isolation: it arises from their connections to other entities and events within a broader context~\cite{ahlqvist2020relational}.

\begin{figure}[t]
\begin{tcolorbox}[
  colback=gray!3, colframe=gray!55, boxrule=0.4pt, arc=2pt,
  left=7pt, right=7pt, top=6pt, bottom=6pt,
  fonttitle=\bfseries\small,
  title={Example: from a rare element to a strong signal}]
{\small
\textbf{\textcolor{orange!70!black}{Rare} (10 January 2024).}
OpenAI quietly removes ``military and warfare'' from the prohibited-uses list in
its usage policy. No announcement. Reported by a single outlet.

\vspace{3pt}\hrule\vspace{4pt}

\textbf{\textcolor{orange!70!black}{Corroboration and potential weak signal}
(16 January 2024).}
At Davos, OpenAI confirms work with DARPA on cybersecurity tools, and states the
blanket ban had blocked legitimate use cases.

\vspace{3pt}\hrule\vspace{4pt}

\textbf{\textcolor{green!35!black}{Strong signal} (16 June 2025).}
A US Department of Defense contract (CDAO), reportedly \$200M, alongside the
launch of ``OpenAI for Government''. The policy edit was individually dismissible; its significance lay in corroboration
}
\end{tcolorbox}
\caption{Weak signals precede a strong signal as fragmentary, low-visibility
traces whose significance emerges only from their corroboration.}
\label{fig:motivating-example}
\end{figure}

Existing methods are primarily keyword frequency-based. They evaluate the distribution of terms across documents to classify them as weak, latent, or strong signals~\cite{yoon2012detecting}. More recent approaches leverage transformer-based topic models within online learning frameworks to classify topics along a noise-to-signal continuum~\cite{boutaleb2024bertrend,ebadi2024wisdom}. However, these methods reduce rich domain knowledge to statistical distributions over topics, failing to capture connections between entities. To incorporate structural dynamics, topological approaches have been proposed. For instance, ~\cite{aboujamra2022detection} introduced a framework that uses "graphlets", small, localized subgraph patterns within temporal interaction networks, as structural signatures to detect weak signals. While this approach effectively models the temporal evolution of network interactions, it operates on simple, untyped graphs. Consequently, it still lacks a semantic layer; it cannot represent named entities, typed relations, or the semantic context of how knowledge evolves.

Temporal Knowledge Graphs (TKGs) offer a natural alternative. A TKG encodes a set of documents as timestamped entity-relation triples, for example (\textit{WHO}, \textit{issued alert on}, \textit{monkeypox}, \textit{2022-05-20}), preserving the semantic and temporal layer that the previous methods discard. A TKG, however, is built from the documents available at one moment, whereas monitoring weak signals means reading a source that keeps changing. This raises two challenges. First, many facts assert no validity period at all, so there is no date to attach to them other than the date they were observed. Second, without a dynamic update mechanism, it would be difficult to track events that appear and disappear over time. A dynamic TKG (DTKG) \cite{lairgi2026atom} addresses both by using dual-time modeling, separating when a fact is observed from its validity period. For instance, an article from January 23, 2020 reporting that a virus had spread to ten countries is stored with the observation time $t_{obs} =$ 2020-01-23 and an empty validity period, since the article states no period of its own and the fact may have been true weeks earlier. Because weak signals appear at a given date and gain substance only as later information corroborates them, a DTKG offers a solid foundation for detecting them.

Despite this potential, applying DTKGs to weak signal detection faces a practical limitation: constructing DTKGs from unstructured text has traditionally required domain-specific pipelines and extensive human annotation~\cite{zhong2023acm}. Recent zero- and few-shot LLM-based methods have substantially reduced this cost~\cite{lairgi2024itext2kg,niu2025tree,lairgi2026atom}. However, no existing method addresses the next step: using a dynamic TKG to detect weak signals. TKG reasoning methods are systematically biased toward frequent patterns and focus on link prediction rather than emergence detection~\cite{cai2023survey,xia2024chain}.
 
In this paper, we propose a framework for weak signal detection in dynamic TKGs. Our contributions are: (i) to the best of our knowledge, this is the first definition of weak signals in dynamic TKGs; (ii) we introduce a self-interpretable framework that operates through two distinct modules: a Rare Subgraphs Extractor, in which each TKG snapshot is processed by an LLM via chain-of-thought (CoT) reasoning to identify subgraphs whose content is in tension with the dominant snapshot narrative, and a Weak Signal Alerter, in which these rare subgraphs are tracked across time steps to isolate weak signals. In the remainder of the paper, we present related work in Section~\ref{sec:related_works}, the proposed approach in Section~\ref{sec:proposed_method}, experimental evaluation in Section~\ref{sec:experiments}, and a conclusion in Section~\ref{sec:conclusion}.

\section{Related works}
\label{sec:related_works}

Early approaches to weak signal detection typically begin by extracting candidate keywords using TF-IDF weighting. These candidates are then evaluated according to two complementary criteria: their occurrence frequency and their temporal growth across successive time windows. Specifically, keywords are considered weak signals when they exhibit both low occurrence frequency, captured by the Degree of Visibility (DoV), and a high growth rate, captured by the Degree of Diffusion (DoD)~\cite{yoon2012detecting}. The resulting weak signals are subsequently projected onto a BCG-style matrix to support expert interpretation. This framework has been replicated across diverse domains~\cite{park2017future,lee2018identification}. However, these keyword-based methods lose the context that gives each term its meaning. 

To address the lack of context limitation of isolated terms, many methods propose topic modeling under the hypothesis that a weak signal manifests as a small cluster that is disjoint from dominant topics. LDA constitutes the predominant technique \cite{maitre2019detection}, extended to the temporal dimension via dynamic topic modeling and topic chain construction, wherein similarity measures track thematic evolution across successive time windows \cite{pepin2017visual,elakrouchi2021endtoend}. More recently, transformer-based topic models such as BERTopic and BERTrend have been proposed to monitor shifts within online learning architectures, continuously classifying topics along a noise-to-signal continuum \cite{boutaleb2024bertrend}. Despite their richer context, all these approaches operate at topic-level granularity, remaining blind to the relational structure that may exist between the entities of the same topic.

Early graph-based methods construct keyword co-occurrence graphs and apply betweenness, degree, and proximity to isolate terms that occupy structurally peripheral yet topologically influential positions as candidate precursors \cite{kwon2018weak,dotsika2017identifying}. Complementary approaches leverage minimum spanning trees to reduce graph complexity while preserving signal-bearing connectivity patterns \cite{kwon2018weak}, and employ 3-cliques and density metrics to characterize early structural anomalies in domain-specific interaction networks \cite{bisson2017strategic}. Moreover, \cite{aboujamra2022beam} proposes BEAM, a framework that enumerates graphlets, all connected induced subgraphs of two to five nodes, across temporally ordered graph snapshots, and characterizes weak signals through the velocity and acceleration of graphlet count series combined with a rarity ratio. While these methods represent a significant advance in structural weak signal detection, they operate on homogeneous, untyped graphs and do not exploit the semantic layer of typed entities and relations.

Current TKG reasoning methods addressing link prediction, such as RE-GCN \cite{li2021temporal}, utilize recurrent graph convolutions; however, their neighborhood aggregation mechanisms tend to dilute sparse interactions, resulting in poor representations of rare entities. While DPCL-Diff \cite{cao20256dpcl} addresses historical sparsity through node diffusion, it assumes new events resemble existing ones, which limits its ability to identify truly novel patterns. LLM-based frameworks like LLM-DA \cite{wang20246llmda} and GenTKG \cite{liao2024gentkg} have introduced rule-guided forecasting, yet they remain biased toward frequent patterns and lack dedicated rarity scoring mechanisms for weak signals. These approaches are optimized for link prediction, leaving them limited in detecting the rare aspect of weak signals.

Despite these advances, current methods face the following limitations: (1) keyword-based approaches capture the weak signal as a term while losing the context that gave each term its meaning. (2) Topic modeling approaches recover that context at the theme level while remaining insensitive to relational structure. (3) Graph-based methods exploit topology but operate on homogeneous, untyped graphs. (4) TKG reasoning frameworks are optimized for link prediction, providing no mechanism to detect weak signals. To address these challenges, we propose a framework for weak signal detection in dynamic TKGs. 

\section{Proposed method}
\label{sec:proposed_method}

\subsection{Definitions}

\paragraph{Dynamic Temporal Knowledge Graph.}
Following \cite{lairgi2026atom}, a Dynamic Temporal Knowledge Graph (DTKG)  is defined as a sequence of TKG snapshots $\{\mathcal{G}_{s}^{t}\}_{t \in \mathcal{T}_{obs}}$, where $\mathcal{T}_{obs}$ is an ordered set of observation timestamps. Each snapshot is defined as:

$$\mathcal{G}_{s}^{t} = \left(\mathcal{E}^{t},\, \mathcal{R}^{t},\,  \mathcal{T}_{start}^{t},\, \mathcal{T}_{end}^{t},\, \mathcal{F}^{t}\right)$$

where $\mathcal{E}^{t}$ is the set of typed entities, $\mathcal{R}^{t}$ the set of typed relations, and $\mathcal{F}^{t}$ the set of quintuples of the form $(e_s, r, e_o, t_{start}, t_{end})$, with $e_s, e_o \in \mathcal{E}^{t}$, $r \in \mathcal{R}^{t}$, and $t_{start} \in \mathcal{T}_{start}^{t}$, $t_{end} \in \mathcal{T}_{end}^{t}$ denoting the validity period of the fact. The dual-time modeling distinguishes between the observation time $t$ and the 
inherent validity period of each fact.

\paragraph{Rare Subgraph.}
A rare subgraph at observation time $t$ is defined as a set of quintuples $\mathcal{S}^{t} \subseteq \mathcal{F}^{t}$ whose content is in tension with the dominant narrative of $\mathcal{G}_{s}^{t}$.

\paragraph{Connecting Subgraph.}
Let $\mathcal{E}(\mathcal{S}^{t})$ denote the set of entities appearing in  the rare subgraph $\mathcal{S}^{t}$. The connecting subgraph $\mathcal{C}^{t}$ is defined as the union of the shortest paths between every pair of entities in $\mathcal{E}(\mathcal{S}^{t})$, computed by breadth-first search over the entity-level adjacency of $\mathcal{G}_{s}^{t}$. $\mathcal{C}^{t}$ preserves the structural context that relates the rare quintuples to one another within the snapshot.

\paragraph{Weak Signal.}
A weak signal is a rare subgraph $\mathcal{S}^{t} \subseteq \mathcal{F}^{t}$ whose content advances a tension already present in a rare subgraph $\mathcal{S}^{t'} \subseteq \mathcal{F}^{t'}$ at a prior observation time $t' < t$. A weak signal is therefore characterized by two properties: rarity with respect to the dominant pattern at time $t$, and corroboration across consecutive snapshots.

Throughout this paper, the term \textit{quintuple} is used as the base unit of the DTKG, where the validity period $(t_{start}, t_{end})$ may or may not be defined depending on whether the underlying fact carries explicit temporal bounds. The DTKG is further considered as a memory in which quintuples carry updatable properties. At each time $t$, the framework operates in two stages.


\subsection{Module 1: Rare Subgraphs Extractor}
\label{sec:rare-elements-extractor}

The TKG snapshot is read from the DTKG. It is then prompted to the LLM, with all quintuples presented as an indexed list of the form (subject entity name: subject entity type $\longrightarrow$ predicate $(t_{start}, t_{end})$ $\longrightarrow$ object entity name: object entity type). Following the CoT technique \cite{wei2022chain}, the LLM reasons in two steps. In the first step, a baseline narrative is established: a summary of the TKG snapshot. In the second step, the quintuples are compared against the baseline narrative to identify rare subgraphs. Deviations arising solely from TKG artefacts, such as label mismatches, that are otherwise aligned with the baseline are explicitly excluded. Once the rare subgraphs have been identified, a connecting subgraph is constructed spanning them. It preserves the structural context that ties the rare subgraphs together while the remainder of the snapshot is discarded. The rare subgraphs and their connecting subgraph at time $t$ are highlighted in memory as prior evidence for the next module.

\subsection{Module 2: Weak Signal Alerter}
\label{sec:weak-signal-alerter}
\begin{figure*}[!t]
  \centering
  \includegraphics[width=\textwidth]{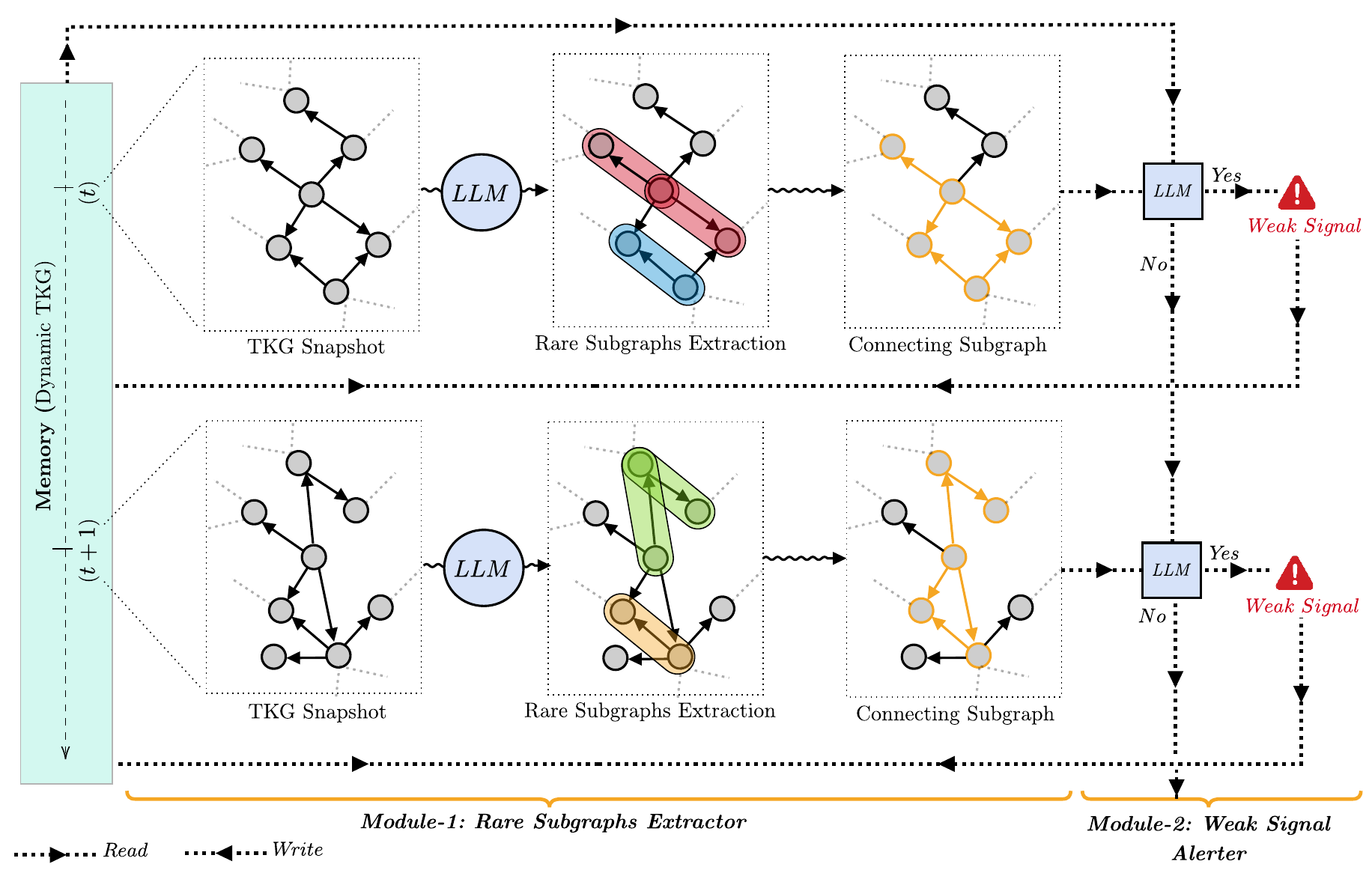}
  \caption{
    \label{fig:architecture}
    The architecture of our framework
  }
\end{figure*}

The connecting subgraph produced by Module~1 is not yet a weak signal. A single deviation observed at time $t$ may fade away or grow into a consequential pattern in later snapshots. The purpose of this module is therefore to decide, by comparing the current connecting subgraph with those of previous snapshots, which subgraph should be labeled as weak signals. The connecting subgraphs of all prior snapshots and the current connecting subgraph $\mathcal{C}^t$ are retrieved from the DTKG. If $\mathcal{C}^t$ is empty, the module is skipped. Otherwise, each subgraph is prompted in the same textual form as in Module~1. Following the CoT technique \cite{wei2022chain}, the LLM reasons in two steps. In the first step, for each past connecting subgraph, the LLM identifies what the snapshot was mainly about and what its rare quintuples were hinting at. In the second step, each quintuple of $\mathcal{C}^t$ is compared against those past hints to decide whether its content continues the same thread. A subgraph is labeled as a \emph{weak signal} if its content advances a tension already present in a prior connecting subgraph, making the earlier rare subgraph reinterpretable as a precursor of a future event. Every labeled subgraph as a weak signal is annotated and written back to memory. At the next time step, this weak signal and the related connecting subgraphs are discarded and are not shown again. Both modules are applied to every subsequent snapshot as shown in Figure~\ref{fig:architecture}.

\section{Experiments}
\label{sec:experiments}

Our evaluation addresses the following research questions:

\noindent\textbf{RQ1:} How accurately does our framework detect weak signals compared to keyword-, topic-, and graph-based methods?

\noindent\textbf{RQ2:} How early are weak signals detected relative to their establishment as strong signals, and how does lead time vary across methods?

\noindent\textbf{RQ3:} Does the framework yield interpretable, contextually grounded explanations supporting its self-interpretable design?

\noindent\textbf{RQ4:} Which components of the framework account for its performance?
\subsection{Metrics}
\label{sec:metrics}
Let $\mathcal{S} = \{s_1, \dots, s_m\}$ denote the set of strong signals. Each signal $s \in \mathcal{S}$ is characterized by two ground-truth timestamps: $t^{\text{weak}}_{s}$, the earliest date at which $s$ qualifies as a weak signal, and $t^{\text{strong}}_{s}$, the date at which it becomes a strong signal. The detection window of $s$ is $\mathcal{W}_s = [\,t^{\text{weak}}_{s},\, t^{\text{strong}}_{s}\,]$. Each method $M$ produces a set of detections $\mathcal{D}_M = \{(\hat{s}, t)\}$, where $\hat{s}$ is the textual or structural description of the detected weak signal and $t$ is its detection timestamp. Each signal $s$ is further associated with a set $\mathcal{A}(s)$ of discriminative anchor words derived from its ground-truth precursor facts.

\paragraph{Detection accuracy (RQ1 and RQ4).}
Because the output representations of the baseline methods differ in granularity: single keywords, bags of topic words, graphlets from which the entities of the underlying graph are collected, and typed quintuples together with a natural-language interpretation, we parameterize matching by a strictness level $k \in \{1, 2, 3\}$. A detection $(\hat{s}, t) \in \mathcal{D}_M$ is a \emph{true positive at level $k$} if (i) the description $\hat{s}$ shares at least $k$ distinct words with $\mathcal{A}(s)$ for some $s \in \mathcal{S}$, and (ii) $t \in \mathcal{W}_s$. Let $\mathcal{S}_M^{+}(k) \subseteq \mathcal{S}$ be the set of strong signals for which $M$ produces at least one true positive at level $k$. Precision, recall, and $F_1$ are reported for each $k$:
\begin{equation}
P_M(k) = \frac{|TP_M(k)|}{|\mathcal{D}_M|},
\qquad
R_M(k) = \frac{|\mathcal{S}_M^{+}(k)|}{|\mathcal{S}|},
\label{eq:precision-recall}
\end{equation}
\begin{equation}
F_{1,M}(k) = \frac{2 \cdot P_M(k) \cdot R_M(k)}{P_M(k) + R_M(k)}.
\label{eq:f1}
\end{equation}
Note that $|\mathcal{D}_M|$ is constant across $k$: stricter matching reduces true positives but does not change how many detections a method emits.

\paragraph{Lead time (RQ2).}
For each signal $s \in \mathcal{S}_M^{+}(k)$, let $t^{\text{det}}_{M,s}(k)$ be the earliest detection timestamp produced by $M$ for $s$ at level $k$. The lead time of $M$ on $s$ is:
\begin{equation}
\mathrm{LeadTime}_M(s,k) = t^{\text{strong}}_{s} - t^{\text{det}}_{M,s}(k).
\label{eq:leadtime}
\end{equation}
We report the mean lead time $\bar{L}_M(k) = \frac{1}{|\mathcal{S}_M^{+}(k)|} \sum_{s \in \mathcal{S}_M^{+}(k)} \mathrm{LeadTime}_M(s,k)$ together with the full per-signal distribution. Lead time captures the early-warning value that $P_M$ and $R_M$ alone cannot: a detection at $t = t^{\text{strong}}_{s} - 1$ is technically correct but operationally useless.

\paragraph{Interpretability (RQ3).}
We provide a qualitative comparison of the outputs produced by each method on the same event.
\subsection{Datasets and baseline methods}
\label{sec:datasets}

No annotated benchmark currently exists for weak signal detection \cite{boutaleb2024bertrend}. State-of-the-art methods are typically evaluated qualitatively by domain experts, and no quantitative gold standard is available for comparison. We introduce what is, to the best of our knowledge, the first benchmark for weak signal detection. 

Constructing such a benchmark presents two main challenges. First, weak signal detection requires access to the complete temporal evolution of a signal. In practice, however, only the resulting strong signal is typically documented, while the weak signals and corroborating events that precede it are rarely recorded. Second, the benchmark must be built from a domain that captures the dynamic and temporal nature of real-world phenomena. We therefore select the corporate domain, as company histories provide heterogeneous, time-stamped event records spanning multiple years.
 
The dataset, referred to as Wiki-OpenAI, is constructed from all monthly edits of the Wikipedia page of OpenAI, collected over the period from January 2015 to December 2025. Atomic facts are extracted from each edit using \texttt{ATOM}~\cite{lairgi2026atom}. Because the same fact may be reproduced across several edits, duplicates are removed. Atomic facts are grouped by the year of the source event, so the observation timestamps of the DTKG are the 11 years 2015-2025. After deduplication, 757 atomic facts are retained from the Wikipedia corpus.

The Wikipedia record does not capture the complete pre-signal history of the strong signals annotated in this benchmark: some facts were present in the real world but absent from the encyclopedic record at the time of their occurrence, and are missing from the snapshots. To address this limitation, 16 atomic facts are added manually. These facts are sourced from established news media (e.g., The New York Times), and the complete list is provided in the GitHub repository. In total, the benchmark contains 773 atomic facts (757 Wikipedia-sourced and 16 manually added).
 
Five strong signals are annotated. Each atomic fact is assigned one of four roles: \textit{background} (no precursor relationship), \textit{weak\_signal} (a pre-signal fact with a documented lead time to a strong signal), \textit{corroboration} (a pre-signal fact that reinforces a prior weak signal), or \textit{strong\_signal} (the anchor event). Post-signal facts are excluded from the scoreable set and are not included in the benchmark. The five strong signals, together with the date of the earliest precursor identified and the date of the strong signal itself, are listed in Table~\ref{tab:strong_signals}. An overview of the full dataset is provided in Table~\ref{tab:dataset_stats}.

The DTKG is constructed from these atomic facts using \texttt{ATOM} \cite{lairgi2026atom}, with the hyperparameters set to their default values. Three baselines are evaluated: Yoon \cite{yoon2012detecting}, BERTrend \cite{boutaleb2024bertrend} and BEAM \cite{aboujamra2022beam}. \texttt{gpt-5.4-mini-2026-03-17}\footnote{\url{https://developers.openai.com/api/docs/models/gpt-5.4-mini}} is chosen thanks to its low cost and performance. Because the benchmark is drawn from a widely documented organization, an LLM may have encountered its history during pretraining. We therefore derive a second variant, \textbf{Wiki-OpenAI-Anon}, in which every named entity is replaced by an anonymized placeholder (\textit{OpenAI} $\rightarrow$ \textit{OrgVrein}, \textit{Pentagon} $\rightarrow$ \textit{GovLaethas}), applying the same substitution to the anchor-word sets $\mathcal{A}(s)$. Graph structure, relation types, and timestamps are unchanged.
\begin{table}[!t]
\centering
\small
\caption{Strong signals annotated in Wiki-OpenAI. The SS date corresponds to $t_\text{est}$, the date on which the event was publicly confirmed.}
\label{tab:strong_signals}
\setlength{\tabcolsep}{4pt}
\begin{tabular}{lp{3cm}cc}
\toprule
\textbf{ID} & \textbf{Description}
            & $t^{\text{weak}}_{s}$
            & $t^{\text{strong}}_{s}$ \\
\midrule
\texttt{SS\_FORPROFIT\_2019}
  & Conversion to capped-profit (OpenAI LP)
  & Dec 2015 & Mar 2019 \\
\texttt{SS\_BOARD\_COUP\_2023}
  & Altman removal and reinstatement
  & Mar 2019 & Nov 2023 \\
\texttt{SS\_NYT\_LAWSUIT\_2023}
  & \textit{NYT} v.\ OpenAI copyright suit
  & Jun 2021 & Dec 2023 \\
\texttt{SS\_DEFENCE\_TURN\_2025}
  & Pivot to defence and national security
  & Jul 2019 & Jun 2025 \\
\texttt{SS\_FORPROFIT\_PBC\_2025}
  & Conversion to for-profit PBC
  & Dec 2017 & Oct 2025 \\
\bottomrule
\end{tabular}
\end{table}

\begin{table}[!t]
\centering
\caption{Summary statistics of the Wiki-OpenAI benchmark dataset.}
\label{tab:dataset_stats}
\begin{tabular}{lc}
\toprule
\textbf{Property} & \textbf{Value} \\
\midrule
\multicolumn{2}{l}{\textit{Corpus}} \\
\quad Source                         & Wikipedia (OpenAI page) \\
\quad Period                & Jan 2015 -- Dec 2025 \\
\quad Number of yearly snapshots            & 11 \\
\midrule
\multicolumn{2}{l}{\textit{Atomic facts}} \\
\quad Total                          & 773 \\
\quad Wikipedia-sourced              & 757 \\
\quad Manually added                 & 16 \\
\midrule
\multicolumn{2}{l}{\textit{Annotation (by role)}} \\
\quad Background                     & 736 \\
\quad Weak signal                    & 22 \\
\quad Corroboration (pre-signal)     & 10 \\
\quad Strong signal                  & 5 \\
\bottomrule
\end{tabular}
\end{table}

\subsection{Results}

\subsubsection{RQ1: Detection Accuracy}

As reported in Table~\ref{tab:detection-accuracy}, \texttt{C-Unseen} obtains the highest $F_{1,M}(k)$ score at $k=2$ and $k=3$, and covers all five strong signals at $k=1$. At $k=1$, BEAM obtains higher precision and $F_1$. All three baselines degrade as $k$ increases: BEAM covers no signal at $k=3$, leaving $F_1$ undefined, while BERTrend retains $1.40 \pm 0.55$ covered signals. Yoon is structurally unable to satisfy higher thresholds, since its detections consist of a single term. The precision and recall of \texttt{C-Unseen} also decrease with $k$, indicating that some detections do not contain sufficient anchor words and therefore constitute false positives; its coverage nevertheless remains at $4.20 \pm 1.10$ signals at $k=3$, against $1.40 \pm 0.55$ for the best baseline. Table~\ref{tab:anonymized} reports the same evaluation on Wiki-OpenAI-Anon. The ranking is unchanged at every threshold, and the degradation of \texttt{C-Unseen} is limited: $F_1$ decreases from $0.613$ to $0.603$ at $k=1$ and from $0.432$ to $0.400$ at $k=2$, with coverage unchanged at five and $4.80 \pm 0.45$ signals respectively. \texttt{C-Unseen} does not detect a weak signal by recognising which company or person it concerns, but by finding facts that conflict with the rest of the snapshot and reappear in later ones. Replacing the entity names leaves both of these intact, which is why performance barely changes. The small residual difference is expected: anonymization removes not only any prior knowledge of the organization, but also the general knowledge that makes a conflict legible. For example, the defence signal rests on the tension between a stated military-use restriction and the interest shown by the Pentagon; once the entity is rendered as \texttt{GovLaethas}, nothing indicates that it is a defence agency, and the conflict is hard to see from the snapshot alone. These results confirm that the proposed framework detects weak signals more accurately than all evaluated baselines, which addresses RQ1.

\begin{table}[!t]
\centering
\caption{Detection accuracy on Wiki-OpenAI at shared-anchor-word match
thresholds $k\in\{1,2,3\}$. \texttt{C-Unseen} and BERTrend are reported as mean\,$\pm$\, sample std over $n{=}5$ runs; Yoon and BEAM are deterministic and reported from a single run. ``--'' denotes an undefined value.}
\label{tab:detection-accuracy}
\small
\setlength{\tabcolsep}{3.7pt}
\begin{tabular}{clccccc}
\toprule
$k$ & Method & $|\mathcal{D}_M|$ & $|\mathcal{S}_M^{+}(k)|$
    & $P_M(k)$ & $R_M(k)$ & $F_{1,M}(k)$ \\
\midrule
\multirow{4}{*}{1}
  & \textbf{BEAM}     & 19 & 3 & \textbf{0.789} & 0.600 & \textbf{0.682} \\
  & \textbf{BERTrend} & 44.8\pmm{5.2} & 4.20\pmm{0.45} & 0.161\pmm{0.021}
                      & 0.840\pmm{0.089} & 0.270\pmm{0.034} \\
  & \textbf{Yoon}     & 311 & 3 & 0.032 & 0.600 & 0.061 \\
  \cmidrule(l){2-7}
  & \textbf{\texttt{C-Unseen}} & 50.2\pmm{10.1} & \textbf{5.00} & 0.444\pmm{0.066}
                      & \textbf{1.000} & 0.613\pmm{0.061} \\
\midrule
\multirow{4}{*}{2}
  & \textbf{BEAM}     & 19 & 1 & 0.053 & 0.200 & 0.083 \\
  & \textbf{BERTrend} & 44.8\pmm{5.2} & 2.80\pmm{1.30} & 0.081\pmm{0.026}
                      & 0.560\pmm{0.261} & 0.141\pmm{0.047} \\
  & \textbf{Yoon}     & 311 & -- & -- & -- & -- \\
  \cmidrule(l){2-7}
  & \textbf{\texttt{C-Unseen}} & 50.2\pmm{10.1} & \textbf{4.80\pmm{0.45}}
                      & \textbf{0.281\pmm{0.048}} & \textbf{0.960\pmm{0.089}}
                      & \textbf{0.432\pmm{0.059}} \\
\midrule
\multirow{4}{*}{3}
  & \textbf{BEAM}     & 19 & 0 & 0.000 & 0.000 & -- \\
  & \textbf{BERTrend} & 44.8\pmm{5.2} & 1.40\pmm{0.55} & 0.043\pmm{0.030}
                      & 0.280\pmm{0.110} & 0.074\pmm{0.050} \\
  & \textbf{Yoon}     & 311 & -- & -- & -- & -- \\
  \cmidrule(l){2-7}
  & \textbf{\texttt{C-Unseen}} & 50.2\pmm{10.1} & \textbf{4.20\pmm{1.10}}
                      & \textbf{0.162\pmm{0.053}} & \textbf{0.840\pmm{0.219}}
                      & \textbf{0.271\pmm{0.084}} \\
\bottomrule
\end{tabular}
\end{table}

\begin{table}[!t]
\centering
\caption{Detection accuracy on \textbf{Wiki-OpenAI-Anon}. Conventions as in Table~\ref{tab:detection-accuracy}.}
\label{tab:anonymized}
\small
\setlength{\tabcolsep}{3.7pt}
\begin{tabular}{clccccc}
\toprule
$k$ & Method & $|\mathcal{D}_M|$ & $|\mathcal{S}_M^{+}(k)|$
    & $P_M(k)$ & $R_M(k)$ & $F_{1,M}(k)$ \\
\midrule
\multirow{4}{*}{1}
  & \textbf{BEAM}     & 20 & 3 & \textbf{0.750} & 0.600 & \textbf{0.667} \\
  & \textbf{BERTrend} & 49.8\pmm{2.0} & 4.00 & 0.120\pmm{0.035} & 0.800
                      & 0.207\pmm{0.053} \\
  & \textbf{Yoon}     & 273 & 3 & 0.022 & 0.600 & 0.042 \\
  \cmidrule(l){2-7}
  & \textbf{\texttt{C-Unseen}} & 58.4\pmm{19.9} & \textbf{5.00} & 0.434\pmm{0.076}
                      & \textbf{1.000} & 0.603\pmm{0.073} \\
\midrule
\multirow{4}{*}{2}
  & \textbf{BEAM}     & 20 & 1 & 0.050 & 0.200 & 0.080 \\
  & \textbf{BERTrend} & 49.8\pmm{2.0} & 3.20\pmm{0.45} & 0.092\pmm{0.028}
                      & 0.640\pmm{0.089} & 0.159\pmm{0.044} \\
  & \textbf{Yoon}     & 273 & -- & -- & -- & -- \\
  \cmidrule(l){2-7}
  & \textbf{\texttt{C-Unseen}} & 58.4\pmm{19.9} & \textbf{4.80\pmm{0.45}}
                      & \textbf{0.257\pmm{0.066}} & \textbf{0.960\pmm{0.089}}
                      & \textbf{0.400\pmm{0.072}} \\
\midrule
\multirow{4}{*}{3}
  & \textbf{BEAM}     & 20 & 0 & 0.000 & 0.000 & -- \\
  & \textbf{BERTrend} & 49.8\pmm{2.0} & 1.80\pmm{0.45} & 0.036\pmm{0.009}
                      & 0.360\pmm{0.089} & 0.066\pmm{0.016} \\
  & \textbf{Yoon}     & 273 & -- & -- & -- & -- \\
  \cmidrule(l){2-7}
  & \textbf{\texttt{C-Unseen}} & 58.4\pmm{19.9} & \textbf{4.20\pmm{1.10}}
                      & \textbf{0.115\pmm{0.057}} & \textbf{0.840\pmm{0.219}}
                      & \textbf{0.196\pmm{0.086}} \\
\bottomrule
\end{tabular}
\end{table}

\subsubsection{RQ2: Lead Time}
As reported in Table~\ref{tab:lead-time}, \texttt{C-Unseen} achieves a mean lead time of $\bar{L}^{\text{years}}_{M}(k=3) = 1.00$ year, compared to $1.70 \pm 0.45$ years for BERTrend. BEAM produces no covered signal at $k=3$, rendering its mean lead time undefined. Yoon is not adapted at $k=3$. This result should, however, be interpreted alongside the recall results in Table~\ref{tab:detection-accuracy}: the lead time of BERTrend is computed over $1.40 \pm 0.55$ covered signals, whereas the lead time of \texttt{C-Unseen} is computed over $4.20 \pm 1.10$. A mean lead time derived from one or two detections may reflect particularly early-detectable signals rather than a consistent early-warning capacity, and a method that covers the remaining signals necessarily includes those detected later. The mean lead time of \texttt{C-Unseen}, combined with its broader signal coverage, therefore indicates that the framework detects weak signals consistently across signals of varying detectability. This result addresses RQ2.

\begin{table}[!t]
\centering
\caption{Mean lead time $\bar{L}_M^{\text{years}}(k{=}3)$ over covered signals. ``--'' denotes no covered signal (mean undefined).}
\label{tab:lead-time}
\small
\setlength{\tabcolsep}{5pt}
\begin{tabular}{lcc}
\toprule
Method & \textbf{Wiki-OpenAI} & \textbf{Wiki-OpenAI-Anon} \\
\midrule
\textbf{BEAM}     & --  & --  \\
\textbf{BERTrend} & 1.70\pmm{0.45} 
                  & 2.00  \\
\textbf{Yoon}     & --  & --  \\
\midrule
\textbf{\texttt{C-Unseen}} & \textbf{1.00}
                  & \textbf{1.13\pmm{0.30}} \\
\bottomrule
\end{tabular}
\end{table}

\subsubsection{RQ3: Interpretability}
\begin{figure*}[!t]
  \centering
  \includegraphics[width=\textwidth]{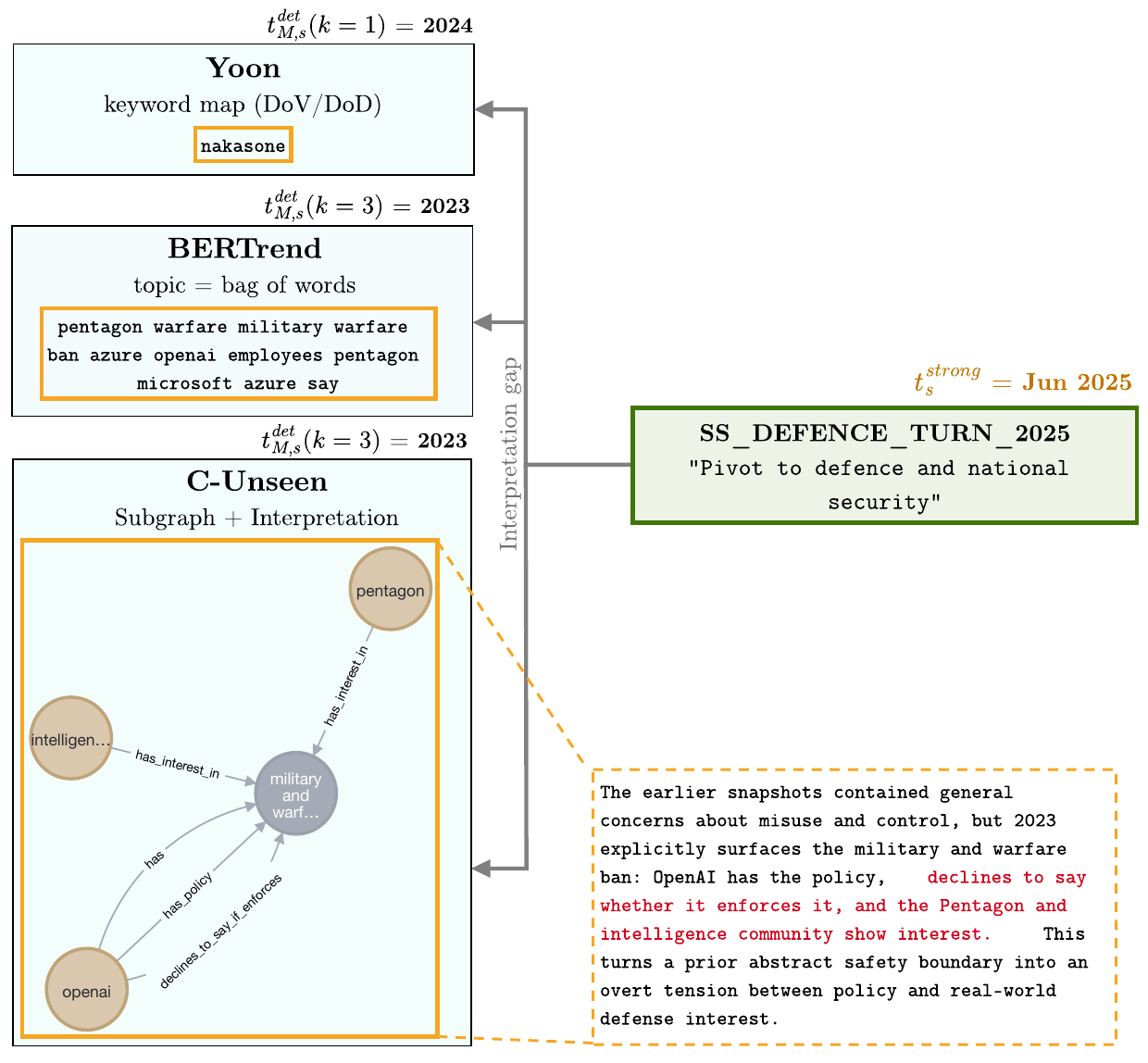}
  \caption{
    \label{fig:qualitative_comparison}
    Qualitative comparison of the baseline methods' output over the detected \texttt{SS\_DEFENCE\_TURN\_2025} by Yoon, BERTrend, and \texttt{C-Unseen}. It was not detected by BEAM.
  }
\end{figure*}

Figure~\ref{fig:qualitative_comparison} provides a qualitative comparison of method outputs for the signal \texttt{SS\_DEFENCE\_TURN\_2025}, which Yoon, BERTrend, and \texttt{C-Unseen} detect but BEAM does not. Yoon identifies a single keyword, which does not provide context for why it constitutes a weak signal related to the defence pivot. BERTrend produces a bag of words that captures a relevant thematic area but does not articulate the relational structure behind the signal nor the reasoning that connects it to the strong signal. In contrast, \texttt{C-Unseen} produces the weak signal as a subgraph of a TKG snapshot, exposing the structural tensions. This subgraph is accompanied by a natural language interpretation that traces the signal's evolution across snapshots, identifying the transition from an abstract safety boundary to an overt tension between policy and real-world defence interests. These results confirm that the proposed framework produces self-interpretable outputs, which answers RQ3.

\subsubsection{RQ4: Ablation}

Table~\ref{tab:ablation} compares four arms against the full framework at $k=3$. Local arms see only the quintuples of the current snapshot; full DTKG arms additionally see those of all prior snapshots. Flat arms receive a flat prompt (e.g., \texttt{detect weak signals in the DTKG quintuples below}), while CoT arms receive the two-step narrative-then-deviation prompt. $\Delta F_1$ is the per-run difference against \texttt{C-Unseen}. CoT reasoning improves on a flat prompt for the local snapshot arms, whereas extending the input to the full DTKG does not add a measurable gain over the corresponding local snapshot arm. Selecting the rare subgraphs and their connecting subgraph instead of prompting the complete graph yields $\Delta F_1 = +0.141 \pm 0.093$ over the full DTKG arm. This addresses RQ4: the gain comes from the combination of the selected rare subgraphs and the CoT prompt, not from the LLM alone.

\begin{table}[!t]
\centering
\caption{Ablation on Wiki-OpenAI at $k{=}3$, mean\,$\pm$\,sample std over $n{=}5$ runs.}
\label{tab:ablation}
\small
\setlength{\tabcolsep}{3pt}
\begin{tabular}{lccccc}
\toprule
Arm & $|\mathcal{D}_M|$ & $|\mathcal{S}_M^{+}|$ & $P_M$ & $F_{1,M}$
    & $\Delta F_1$ \\
\midrule
local+flat  & 144.8\pmm{18.5} & 5.00 & 0.059\pmm{0.013} & 0.112\pmm{0.023}
            & $-$0.159 \\
local+CoT   & 135.2\pmm{19.9} & 4.80\pmm{0.45} & 0.072\pmm{0.010}
            & 0.134\pmm{0.018} & $-$0.137\pmm{0.089} \\
full DTKG+flat & 195.0\pmm{14.1} & 5.00 & 0.061\pmm{0.009} & 0.115\pmm{0.016}
            & $-$0.156 \\
full DTKG+CoT  & 189.6\pmm{10.5} & 5.00 & 0.069\pmm{0.013} & 0.130\pmm{0.023}
            & $-$0.141\pmm{0.093} \\
\midrule
\textbf{\texttt{C-Unseen}} & \textbf{50.2\pmm{10.1}} & 4.20\pmm{1.10}
            & \textbf{0.162\pmm{0.053}} & \textbf{0.271\pmm{0.084}} & --- \\
\bottomrule
\end{tabular}
\end{table}

\section{Conclusion}
\label{sec:conclusion}
In this paper, we introduced \texttt{C-Unseen}, a self-interpretable framework for weak signal detection in Dynamic Temporal Knowledge Graphs, along with the formal definition of weak signals in this setting. The DTKG served three complementary roles in the framework: as a storage of domain knowledge over time, as a persistent memory in which the intermediate reasoning steps of the LLM are stored across snapshots, and as a structural backbone that expands the context available to the Weak Signal Alerter by connecting rare subgraphs. Experimental evaluation on the Wiki-OpenAI benchmark demonstrated that the framework achieves better performance, with self-interpretable outputs, and that this performance is
retained on a named entity anonymized variant of the benchmark. Several directions are identified for future work. First, the scalability of the Rare Subgraphs Extractor to large DTKGs remains to be addressed: as the number of quintuples per snapshot grows, the full snapshot may exceed the LLM's context window. A direction is to prompt the model over a compressed representation of the TKG snapshot, guided toward subgraph regions that are likely to contain rare subgraphs, rather than presenting the complete snapshot. Second, the DTKG in this work is constructed in a few-shot manner using \texttt{ATOM}; the introduction of a domain ontology to guide \texttt{ATOM} during construction is expected to reduce false positives. Third, the benchmark introduced here is limited to a single organization; evaluation across multiple domains would enhance the generalizability of the framework.

%
%
%
\bibliographystyle{splncs04}
%
\bibliography{sample}

\end{document}